\documentclass[lettersize,journal]{IEEEtran}
\usepackage{amsmath,amsfonts}
\usepackage{algorithmic}
\usepackage{algorithm}
\usepackage{array}
\usepackage{ragged2e}
\usepackage[caption=false,font=normalsize,labelfont=sf,textfont=sf]{subfig}
\usepackage{textcomp}
\usepackage{stfloats}
\usepackage{url}
\usepackage{verbatim}
\usepackage{graphicx}
\usepackage{cite}

\usepackage{amsmath}
\usepackage{times}
\usepackage{latexsym}
\usepackage{multirow}
\usepackage{wrapfig,lipsum,booktabs}
\usepackage{caption}
\usepackage{subcaption}
\usepackage{microtype}
\usepackage[numbers]{natbib}

\emergencystretch=\maxdimen
\begin{document}

\title{\textit{LineConGraphs}: Line Conversation Graphs for Effective Emotion Recognition  using Graph Neural Networks}
%Gokul S Krishnan, Sarala Padi, Craig S. Greenberg, Balaraman Ravindran, Dinesh Manoch and Ram D.Sriram
\author{\RaggedRight{Gokul S Krishnan, Centre for Responsible AI, IIT Madras, Chennai, India}\\
\RaggedRight{Sarala Padi$^{*}$, National Institute of Standards and Technology, Gaithersburg, Maryland, USA}\\
Craig S. Greenberg, National Institute of Standards and Technology, Gaithersburg, Maryland, USA \\
Balaraman Ravindran, 
Computer Science \& Engineering, Centre for Responsible AI, IIT Madras, Chennai, India\\
Dinesh ~Manocha, Computer Science \& Engineering, University of Maryland, College Park, Maryland, USA\\
Ram ~D. Sriram, National Institute of Standards and Technology, Gaithersburg, Maryland, USA\\
~~~~~~${*}$ Corresponding author: {sarala.padi@nist.gov}}
        % <-this % stops a space
%\thanks{This paper was produced by the IEEE Publication Technology Group. They are in Piscataway, NJ.}% <-this % stops a space
%\thanks{Manuscript received April 19, 2021; revised August 16, 2021.}}

% The paper headers
%\markboth{Journal of \LaTeX\ Class Files,~Vol.~14, No.~8, August~2021}%
%{Shell \MakeLowercase{\textit{et al.}}: A Sample Article Using IEEEtran.cls for IEEE Journals}

%\IEEEpubid{0000--0000/00\$00.00~\copyright~2021 IEEE}
% Remember, if you use this you must call \IEEEpubidadjcol in the second
% column for its text to clear the IEEEpubid mark.

\maketitle

\begin{abstract}

% Modified on 30th Nov
Emotion Recognition in Conversations (ERC) is a critical aspect of affective computing, and it has many practical applications in healthcare, education, chatbots, and social media platforms. Earlier approaches for ERC analysis involved modeling both speaker and long-term contextual information using graph neural network architectures. However, it is ideal to deploy speaker-independent models for real-world applications. Additionally, long context windows can potentially create confusion in recognizing the emotion of an utterance in a conversation. To overcome these limitations, we propose novel line conversation graph convolutional network (\textit{LineConGCN}) and graph attention (\textit{LineConGAT}) models for ERC analysis. These models are speaker-independent and built using a graph construction strategy for conversations -- line conversation graphs (\textit{LineConGraphs}). The conversational context in \textit{LineConGraphs}  is short-term -- limited to one previous and future utterance, and speaker information is not part of the graph. We evaluate the performance of our proposed models on two benchmark datasets, $\textrm{IEMOCAP}$ and $\textrm{MELD}$, and show that our \textit{LineConGAT} model outperforms the state-of-the-art methods with an F1-score of $\textrm{64.58\%}$ and $\textrm{76.50\%}$. Moreover, we demonstrate that embedding sentiment shift information into line conversation graphs further enhances the ERC performance in the case of GCN models.

\end{abstract}


\section{Introduction} \label{intro}

Automatic Emotion recognition (ER) is an essential tool in human-computer interaction (HCI) as it allows people to interact with robots more naturally ~\cite{EmotionHCICowie2001,EmotionIntroSchuller2018}. It also plays a significant role in various sectors, such as healthcare, education, surveillance, security, and automation~\cite{HCIHarper2008}. Emotion Recognition in Conversation (ERC) is a sub-field of ER that involves identifying the emotions of speakers in a given conversation as opposed to utterance level. Humans express emotions in various ways, resulting in different modalities of data such as video, audio, and text that can be potentially used for ERC. However, in the modern world, there has been a significant shift in the way textual content is created and used, enabling us to collect and mine textual data to analyze people's emotions. This shift has opened up new avenues for ERC analysis, and the analysis of textual data is now an important part of the ERC study. So, in this study, we focus on text utterances at a conversational level for ERC analysis.  

In a conversation, speakers tend to maintain a stable emotional trend in line with their speaking logic, which is known as emotional inertia. This phenomenon has been discussed by Poria et al.~\cite{ERCChallengesPoria2019}. However, conversations have a unique nature, and speakers' emotions can be influenced by other speakers, resulting in emotion shifts~\cite{DialogXLShen2020}. Previous models have primarily focused on emotional inertia and paid less attention to emotion shifts, making emotion recognition errors more likely to occur during emotion shifts~\cite{ERCChallengesPoria2019}. To tackle this challenge, researchers have proposed several models, including multi-talk learning models~\cite{ESDERCGao2022, MultitaskSchuller2022} and crossmodal fusion networks \cite{SpeakerIndSchuller2022,EmoBedSchuller2019} with emotion-shift awareness~\cite{CFN-ESALi2023}. Gao et al.~\cite{ESDERCGao2022} proposed a multi-task learning model for emotion recognition that exploits emotion shift detection. However, the model's dependence on speaker information to generate emotion shift labels limited the effectiveness of the approach. On the other hand, Jiang Li et al.~\cite{CFN-ESALi2023} proposed a Cross-modal Fusion Network with Emotion-Shift Awareness (CFN-ESA) for emotion recognition. The model concatenates utterance level and emotion shift features for audio, video, and text modalities and utilizes a multi-head attention recurrent neural network to optimize for ERC and emotion shift classification tasks. While both methods generated emotion shift labels from the actual emotional labels, it is not evident from the reported results that integrating emotion shift significantly improved the ERC performance. One potential reason for this could be that emotions and emotion shifts generated from emotion labels do not convey distinct information to each other.

Individuals express emotions in a unique way that can differ from person to person. To improve emotional recognition and understanding in conversations, it may be helpful to consider speaker-specific modeling based on preceding utterances. There are several proposed models that capture the emotional trends and dynamics of participants, including DialogXL\cite{DialogXLShen2020}, CauAIN~\cite{ CauAINZhao2022}, SAPBERT~\cite{SAPBERTLim2023}, DialogueEIN~\cite{DialogEINLiu2022}, and RBAGCN~\cite{RBA-GCNLin2023}. Additionally, several graph-based models, such as MMGCN~\cite{hu2021mmgcn}, RobERTa~\cite{kim2021emoberta},  Interactive Conversational memory Network (ICON)~\cite{ICONHazarika2018}, conversational GCN~\cite{ConGCNZhang2019}, DialogRNN~\cite{majumder2019dialoguernn}, DialogGCN~\cite{ghosal2019dialoguegcn}, have utilized speaker information to model inter-intra speaker relationships and improve ERC performance. These models use various approaches to model inter and intra-speaker dependencies, causal-aware interaction networks, speaker identification, emotional interaction networks, static and dynamic states of speakers~\cite{StaticDynamicSaxena2022}, and relational bilevel aggregation graph convolutional networks, to enhance emotional recognition accuracy in conversations. However, it's important to note that it's not always clear whether speaker modeling improves emotion recognition performance in ERC analysis. Results have shown that including speakers without any random speaker initializations, the performance is almost similar. This suggests that embedding or modeling speaker information into graph neural networks may not always improve model performance and may even create noise by feeding too much context, which could affect the ERC performance.

%Context is crucial in natural language processing (NLP) research domain, and contextual embeddings have been shown to significantly enhance the performance of NLP systems~\cite{ SAPBERTLim2023}. In addition,  context plays a significant role in recognizing the emotions of utterances in each conversation. The relevance of context varies depending on the problem and can be derived from local or distant conversational history. While local context is more apparent, distant context can also be useful when a speaker refers to earlier utterances spoken by anyone in the conversational history~\cite{ContextERCWang2020}. In conversations, determining the emotion of an utterance at a specific time can be done by considering the preceding utterances at a time less than the current time as its context. However, computing this context and deciding how much of it to consider can be challenging due to the dynamic nature of emotions. Several methods have been developed to capture long-range global context, including contextual attention networks~\cite{AfCANWang2021}, context and sentiment-aware frameworks~\cite{SenticGATGeng2022}, Dual stream Recurrence attention networks~\cite{DualRANLi2023}, and conditional random fields~\cite{ContextERCWang2020}, to improve ERC performance.  LIMITATION OF PRIOR METHODS AND WHAT WE ARE DOING HERE.

Context is crucial in natural language processing (NLP) research domain, and contextual embeddings have been shown to significantly enhance the performance of NLP systems~\cite{SAPBERTLim2023}. In addition,  context plays a significant role in recognizing the emotions of utterances in each conversation. The relevance of context varies depending on the problem and can be derived from local or distant conversational history. While local context is more apparent, distant context can also be useful when a speaker refers to earlier utterances spoken by anyone in the conversational history~\cite{ContextERCWang2020}. In conversations, determining the emotion of an utterance at a specific time can be done by considering the preceding utterances at a time less than the current time as its context.  Several methods have been developed to capture long-range global context, including contextual attention networks~\cite{AfCANWang2021}, context and sentiment-aware frameworks~\cite{SenticGATGeng2022}, Dual stream Recurrence attention networks~\cite{DualRANLi2023}, and conditional random fields~\cite{ContextERCWang2020}, to improve ERC performance.  However, computing this context and deciding how much of it to consider can be challenging due to the dynamic nature of emotions.

As mentioned earlier, prior methods of ERC analysis incorporated speaker information, emotion shift, and contextual information using graph neural network architectures. However, these approaches have limitations due to the unavailability of the speaker's identity or impracticality in building and deploying speaker-independent models for real-time applications. Moreover, recognizing the emotion of an utterance in a conversation can become confusing due to long-distance context. To address these limitations, we propose two novel models for ERC analysis: the line conversation graph convolutional network (\textit{LineConGCN}) and the graph attention (\textit{LineConGAT}) models, developed on line conversation graphs (\textit{LineConGraphs}) by utilizing node feature representations extracted from a transformer model. These models are speaker-independent, and the context in \textit{LineConGraphs} is limited to one previous and future utterance. We evaluated these models on two benchmark datasets, $\textrm{IEMOCAP}$ and $\textrm{MELD}$, and demonstrated that our \textit{LineConGAT} model outperforms the state-of-the-art methods. Furthermore, we showed that embedding sentiment shift information into \textit{LineConGCN} can further enhance the ERC performance.

%To address these limitations, we propose a new approach called \textit{LineConGraphs} , which represents conversations such that the emotion recognition models consider only the context of a particular conversation under consideration. We make use of Graph Neural Networks (GNNs) -- Graph Convolution Networks (GCN) and Graph Attention (GAT) models, to effectively learn the graphs constructed using the proposed representation strategy. The proposed approach has been evaluated on two benchmark datasets -- IEMOCAP and MELD and has shown superior performance when compared to other methods. 

The main contributions of this paper are follows:
\begin{enumerate}
    \item Propose Line Conversation Graphs (\textit{{LineConGraphs}}), a speaker independent approach to represent conversations in the form of graphs, utilizing node feature representations extracted from transformer models.
    \item Design LineConGCN and LinConGAT models to learn the nuances of the conversations represented by the proposed \textit{{LineConGraph}} approach.
    \item Embed sentiment shift information into \textit{LineConGraphs} to capture the change in emotions for ERC analysis.
    \item Compare the proposed approach to the state-of-the-art (SOTA) methods for ERC using two benchmark datasets, $\textrm{IEMOCAP}$ and $\textrm{MELD}$. We also look at the proposed  approach in greater detail.
   
    %Furthermore, we compare the performance of the proposed approach against the state-of-the-art (SOTA) methods for ERC using two benchmark datasets --  $\textrm{IEMOCAP}$ and  $\textrm{MELD}$, and we also examine the efficacy of the proposed \textit{{LineConGraphs}} approach in greater detail.

\end{enumerate}

The rest of the paper is organized as follows: Section~\ref{related_work} delves into previous methods of constructing the graph representation for given conversations and developing graph neural network-based models for ERC analysis. Section~\ref{methods} goes into detail about the proposed methodology for constructing the graph  and applying GNNs for ERC. Sections~\ref{datasets} and ~\ref{results} provide a comprehensive discussion of the dataset utilized for the ERC analysis and provide a comprehensive analysis of the experiment results. Finally, Section~\ref{conclusion} concludes the paper with prospective future scope.

%===============================================================================================================
\section{Related Work} \label{related_work}

When it comes to analyzing context-based utterances for ERC analysis, there are several successful methods that have been applied. Graph-based neural networks and transformer-based models have both been used to model long-term dependencies between words in utterances and relationships between utterances~\cite{GNNSurveyWu}. However, prior methods have mostly focused on modeling the context extending the graph-based neural networks to model utterances, speakers, and relationships in a conversation~\cite{HCIHarper2008}. Emotional dynamics in conversations play a crucial role in ERC analysis, including inter- and intra-personal dependencies~\cite{MORRIS20001}. To model these interactions, researchers have proposed several frameworks. For instance, \citet{ICONHazarika2018} proposed the Interactive COnversational memory Network (ICON), which is a multimodal emotion detection framework that hierarchically models the self- and inter-speaker emotional influences into a global context. Similarly, \citet{ConGCNZhang2019} proposed a conversational graph convolutional neural network (ConGCN) that models utterances and speakers in a given conversation. Another approach is DialogRNN~\cite{majumder2019dialoguernn}, which builds recurrent neural network models by jointly encoding the preceding utterances and speaker states for context representation along with emotion representation from the state and preceding speaker states as context. Dialog graph convolutional networks (DialogGCN)~\cite{ghosal2019dialoguegcn} is another model that captures the long-distance context in given conversations. 

\citet{hu2021mmgcn} proposed a multimodal fused GCN (MMGCN) that effectively utilizes both multimodal and long-distance contextual information by leveraging speaker information to capture inter- and intra-speaker dependencies. Similarly, \citet{kim2021emoberta} proposed  EmoBERTa approach which can learn intra- and inter-speaker states and context to predict the emotion of a current speaker by simply prepending speaker names to utterances and inserting separation tokens between the utterances in a dialogue. \citet{RGCNChoi2021} proposed a residual-based graph convolution network (RGCN),  to fully exploit the intra–inter informative features. This approach achieves superior performance compared with state-of-the-art methods, demonstrating the importance of explicit speaker modeling. \citet{StaticDynamicSaxena2022} recent study considers both conversational context and speaker personality by building GNN encoder to model the internal state of the speaker (personality) as a Static and Dynamic speaker state and reported state-of-the-art results.  However, by embedding or modeling speaker information into GNN model does not always improve the model performance, and can instead create a considerable amount of noise by feeding too much context. 

\citet{SkierWei2023} introduced an ERC model that considers discourse relations and symbolic knowledge in multi-party conversations by generating the context-aware transformer representations and knowledge extraction to facilitate a better understanding of an emotion in a given conversations.  Similarly, \citet{ClusterContrastiveYang2023} proposed Supervised Cluster-level Contrastive Learning (SCCL) to enhance the encoder for extracting context-aware representations by utilizing pre-trained knowledge adapters for ERC. On the other hand, \cite{zhang2023multitask,MultitaskSchuller2022}  proposed a multimodal multitask learning models  using intramodal and intermodal attention mechanisms for  emotion recognition analysis. Furthermore, \citet{JiangLi2023Neuro} proposed a multimodal fusion-based approach called GraphMFT. The approach uses an improved Graph Attention (GAT) model to extract representations from three graphs consisting of text, vision, and audio modalities. Additionally, the approach utilizes speaker embeddings similar to MMGCN~\cite{hu2021mmgcn} to enhance the predictions of emotions, which makes it speaker-dependent.

%\citet{SkierWei2023} proposed addressing the task of ERC by considering discourse relations between utterances and incorporating symbolic knowledge into multi-party conversations. They developed specific modules for generating dialog graphs of conversations, generating context-aware transformer representations, knowledge extraction to identify casualties of contexts and used these along with a symbolic fusion attention module to identify emotions.
%\cite{zhang2023multitask} propose a multimodal multitask learning model based on the encoder–decoder architecture, termed M2Seq2Seq, that consists of an intramodal and intermodal attention mechanisms to model contextual dependencies and multimodal relationships of utterances. They also construct single-level and multi-level multitask decoders for performing sarcasm, sentiment and emotion recognition in conversations.

%-------------------------------------------------------------------------------------------------------------------------------------
\subsection{Graph Construction Strategies}
In this section, we focus on the graph construction strategies of some of the prior approaches for ERC. MMGCN~\cite{hu2021mmgcn} constructs a graph by considering utterances as nodes and connecting nodes from different modalities, such as audio, text, and video, based on the corresponding utterances. This allows for the modeling of contextual information across multiple modalities. Interestingly, the presence of speaker information in the MMGCN network did not significantly enhance emotion recognition performance in conversations, as indicated by the nearly identical F1 score with and without speaker embedding across various datasets. Other models, such as ConGCN~\cite{ConGCNZhang2019} and GraphMFT~\cite{JiangLi2023Neuro} also use graph construction for emotion recognition analysis. The former constructs a graph by taking into account the utterances and speakers as nodes, while the latter considers the utterances as nodes and connects all past and future utterances in a conversation to the node. GraphMFT also considers two modalities at a time to represent the graph and includes speaker embeddings similar to MMGCN to slightly improve the ERC performance. 

\citet{StaticDynamicSaxena2022} created a graph using utterance and speaker information as nodes. The graph was fully connected but limited to a context of three utterances (previous and next) for GCN and five for GAT models. The results showed that speaker embedding did not significantly improve model performance, with a difference of only 1.2\%. The authors only considered the past utterance context for emotion recognition analysis. Similarly, \citet{ghosal2019dialoguegcn} hypothesized that each utterance in a conversation is contextually dependent on all other utterances, but constructing a fully connected graph is computationally expensive. Instead, the authors used a practical solution of constructing edges with a past and future context window size of ten. \citet{RGCNChoi2021} also used a fully connected graph with a fixed window size of ten, without speaker information. Modeling the attention weights improved the F1 score for the  $\textrm{IEMOCAP}$ dataset, but not for the  $\textrm{MELD}$ dataset. Additionally, it is not clear how long-distance context or speaker information can help improve the emotional recognition of utterances. Further, extending the context window beyond a conversation could aggregate too much information and make it more challenging for the model to identify emotions. Overall, while graph-based approaches have shown promise in emotion recognition in conversations, the role of speaker information and long-distance context is still unclear. In this paper, we focus on building a speaker-independent model that considers a short context window of previous and next utterances for effectively modeling emotional categories for ERC.

%================================================================================================================
\section{Methodology}\label{methods}

\begin{figure*}[h!]
    \centering
    \includegraphics[width=17cm, height=5cm]{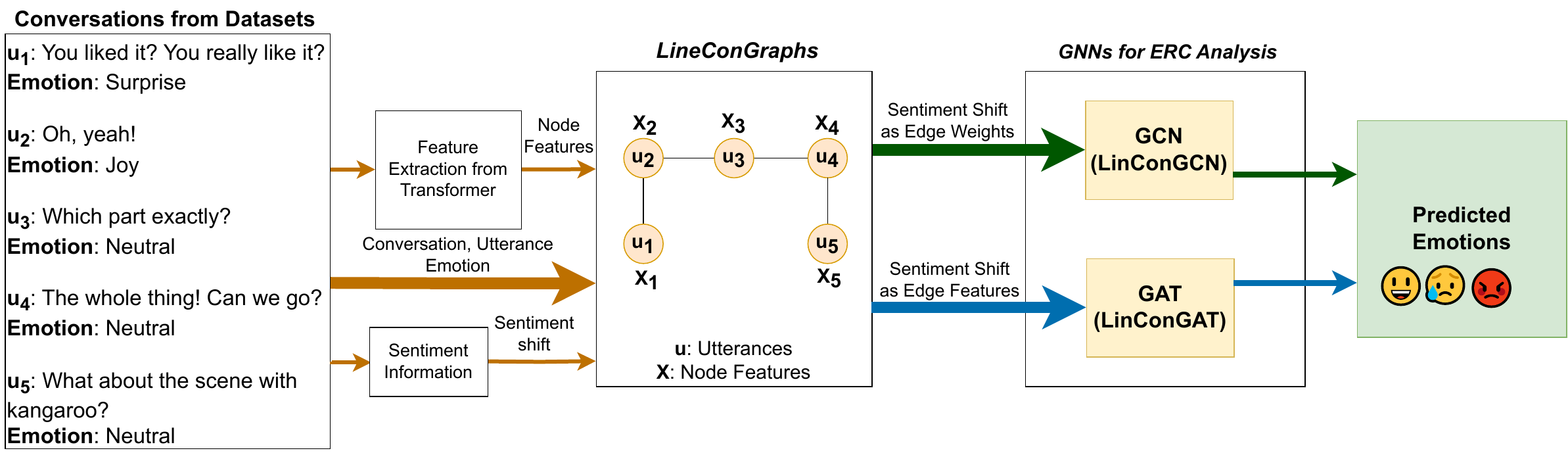}
    \caption{The building blocks of the proposed framework for ERC analysis. }
    \label{fig:overall}

\end{figure*}

The Figure~\ref{fig:overall} provides an overview of the proposed methodology for ERC analysis. The framework comprises two main components: graph construction and building the graph neural network models for ERC analysis. The building blocks of the proposed framework for ERC analysis are divided into three parts. The left block contains examples of conversations from the MELD dataset. The middle block describes the line graph construction strategy, while the final block is focused on modeling utterances at the conversation level for emotion recognition analysis. These building blocks form the core of the proposed methodology and are discussed in greater detail in subsequent sections. 

\subsection{\textit{\textit{LineConGraphs}}: Line Conversation Graphs}
Graphs are an immensely powerful tool for describing complex systems. They consist of nodes or objects, connected by edges or interactions. The real strength of graph formalism is its ability to emphasize the relationships between nodes while maintaining a remarkable level of versatility~\cite{GNNBookWilliam}. Recently, graph-based neural networks have been proposed for ERC analysis, demonstrating the immense potential of using graphs for data analysis~\cite{GNNReviewZhou2020, GNNSurveyWu2020, DLGraphsZhang2018}. The graph representation that is fed to the graph-based neural network models plays a critical role in the overall ERC pipeline and its analyses and therefore, we aim to improve the ERC performance by improving the effectiveness of the graph representation.

\begin{figure}[h!]
%\begin{wrapfigure}{r}{0.4\textwidth}
    \centering
    \includegraphics[width=0.35\textwidth]{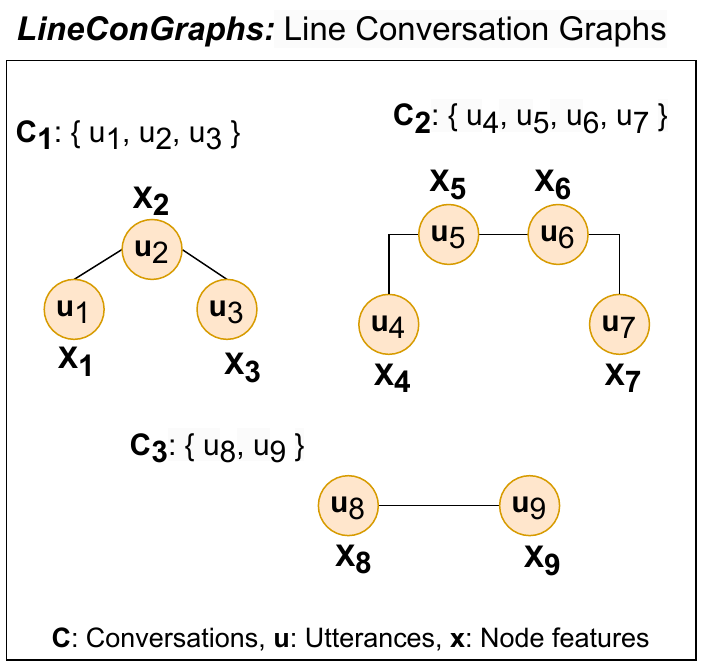}
    \caption{The construction of a line conversation graph for a corpus of conversations. Abbreviations: $C_{i}$: $i^{th}$ conversation; $u_{i}$: $i^{th}$ utterance; $x_{i}$:  feature extracted for $i^{th}$ utterance node; NOTE: self-loops are not shown in the figure for readability}
    \label{fig:graph-construct}
%\end{wrapfigure}
\end{figure}

In a conversation, current utterance emotion mostly depends on the emotions of other utterances, especially the previous ones~\cite{ghosal2019dialoguegcn, navarretta2016mirroring}. This phenomenon is called ``emotion shift'', and we can capture this change in emotion using graphs that consider specific contexts of a conversation. Our proposed approach, called \textit{{LineConGraphs}}, leverages line graphs to identify emotions accurately. Each node in the graph represents a single utterance, and each graph represents an entire conversation. By focusing on the previous and next utterances, we gain invaluable insight into the nuances of ``emotion shift'', allowing our model to effectively identify the emotion of a given utterance in a conversation. What sets \textit{{LineConGraphs}} apart is its ability to consider each conversation in a corpus as a separate entity, ensuring that emotion recognition is performed based solely on the short context of that particular conversation. This approach eliminates the risk of unnecessary context or information creeping in from other conversations in the corpus. We can represent a corpus with many conversations as a single graph of connected utterance nodes, enabling GNN models to generalize learning from patterns and accurately classify similar emotions across different conversations.

Figure~\ref{fig:graph-construct} illustrates how the proposed \textit{{LineConGraphs}} works for given utterances in a conversation. For example, consider a corpus with  three conversations-- $C_{1}$, $C_{2}$ and $C_{3}$,  and nine utterances in total where $C_{1}$ consists of $\{u_{1}, u_{2}, u_{3}\}$, $C_{2}$ contains $\{u_{4}, u_{5}, u_{6}, u_{7}\}$ and $\{u_{8}, u_{9}\}$ are part of $C_{3}$. As we can see in the graph depicted in Figure~\ref{fig:graph-construct}, each utterance is represented by a node, and these nodes are linked by connecting each utterance node to its previous and subsequent nodes. Additionally, we also have self-loops for each node to ensure that the current node becomes a part of the aggregation process in GNNs. In this particular line graph, every node has a degree of 2 (3, including the self-loop)\footnote{NOTE: self-loops are not shown in the Figure~\ref{fig:graph-construct} for readability}, with the exception of the first and last nodes in a conversation, which has a degree of 1 (2 including the self-loop). We generate a graph for every conversation within the specified dataset. As each node represents an utterance, we perform a node classification task on top of the graph to obtain the emotional labels for each utterance. This can be described as:

Given a conversation $C$ with $t$ utterances, $C = { (u_{1} , u_{2} , u_{3} , ... u_{t} })$,  spoken by $m$ speakers $S_{p} = { (sp_{1} , sp_{2} , sp_{3} , ... sp_{m} )}$, with $n$ emotion labels, $E = { (e_{1} , e_{2} , e_{3} , ... e_{t} })$, where each utterance $u \in C$, associated with a speaker $s \in S$, has a labeled emotion $e \in E$. The goal is  to design a model to learn a function   $f (u,C) \rightarrow E$, that can predict the emotion of an utterance $u$ in a given conversation.

%-------------------------------------------------------------------------------------------------------------------------------------

\subsubsection{Node Feature Extraction}

The graph neural network model has a significant advantage in its ability to incorporate node features into the learning process. It has been found that GNNs perform well when there is a strong correlation between node features and node labels~\cite{NodefeatChi2019}. Different ways of initializing node features include a centrality-based approach and a learning-based approach. The former method constructs a node feature based on its local neighborhood, while the latter considers a node feature as the node embedding. Recently,  EmoBERTa~\citep{kim2021emoberta} has been proposed for the task of ERC, which can be used as a pre-trained model for generating effective feature representations of utterances. The proposed \textit{LineConGraphs} approach uses EmoBERTa-base\footnote{\url{https://huggingface.co/tae898/emoberta-base}} for extracting the feature representation of the utterances, which is used as node level feature  that the utterance is associated with. Formally, for each utterance, say $u$, we extract features from the pretrained model associated with the utterance, say $u_{v} \in R^{[1 \times n]}$; where $u_{v}$ is a vector representation of utterance $u$ and $R^{[1 \times n]}$ indicates a vector of real numbers of size $n$, where $n$ is the dimension of the vector extracted for $u$ from the transformer model.

%-------------------------------------------------------------------------------------------------------------------------------------

\subsubsection{Edge Attributes}

\begin{figure*}[hbtp!]
    \centering
    \includegraphics[width=17cm, height=4.5cm]{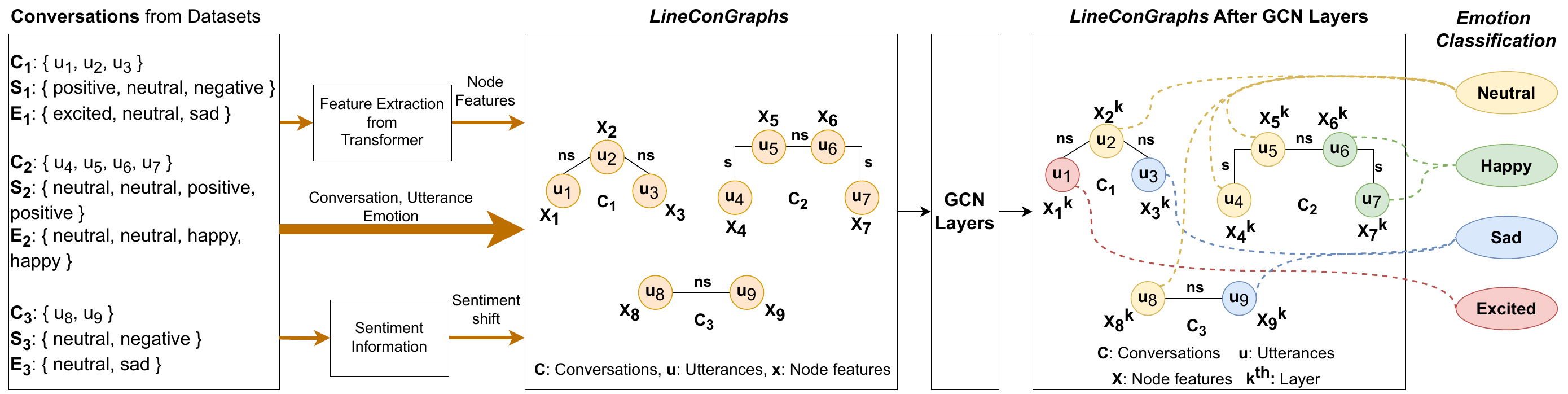}
    \caption{Overview of the proposed framework using the \textit{{LineConGraphs}} with GCN layers for ERC analysis. We use sentiment shift as edge weights where ``$s$'' represents sentiment shift, and  ``$ns$'' indicates there is no sentiment shift. [ $u_{i}$: $i^{th}$ utterance ; $X_{i}$: $i^{th}$ node feature extracted for an utterance; $X_{i}^{k}$ indicates feature representation of $i^{th}$ utterance after $k$ GCN layers]}
    \label{fig:gcn-layers-diag}
\end{figure*}

To better detect emotion shifts in utterances, we can embed sentiment information as edge weights or edge features in our line graph construction process and we refer to this phenomenon as ``sentiment shift''. For GCN models, we use edge weights to represent sentiment shifts, while for GAT models, we directly provide sentiment information as edge features. We assume that this can help the model effectively recognize the emotions of utterances. For example, in the case of GCN,  $u_{i}$ and $u_{j}$ are two connected nodes with an edge weight ``$s$'' if there is a change in the sentiment, otherwise with an edge weight ``$ns$’’ ($s$ and $ns$ indicate `shift' and `no shift' respectively). Similarly, in the case of GAT, the sentiment labels of nodes of an edge is embed as a vector: $[ s_{i}, s_{j} ]$ ($s_{i}$ and $s_{j}$ indicate the sentiment labels of nodes $i$ and $j$ respectively). Figures~\ref{fig:gcn-layers-diag} and \ref{fig:gat-layers-diag} demonstrate the embedded sentiment information in the graphs and the overall learning of GCN and GAT layers for ERC analysis. 

%-------------------------------------------------------------------------------------------------------------------------------------

\subsection{Learning Approaches for ERC Task}

GNNs are powerful tools for representing data relationships in various scientific and engineering fields, such as computer vision, molecular chemistry, molecular biology, pattern recognition, and data mining~\cite{GNNsReviewZhou2021,GNNsSurveyZhang2020}. GNNs combine the strengths of recursive neural networks and random walk models to handle a broad range of graphs, including cyclic, directed, and undirected graphs without any preprocessing~\cite{GNNsScarselli2009,GNNBookWilliam}. The key to GNNs is the generation of node representations, which depend on the graph's structure and feature information encoded at the graph, edge, and node levels. This flexible framework allows for the development of deep neural networks on graph data, overcoming the challenges of complex encoders for graph-structured data. In this paper, we use GNNs to effectively identify emotions in utterances through our \textit{LineConGraphs} approach. We designed two different GNNs, Graph Convolution Network and Graph Attention Network, to achieve this goal. The following sections will elaborate on these models in detail.

%-------------------------------------------------------------------------------------------------------------------------------------

\subsubsection{Graph Convolution Networks}

Graph convolutional networks (GCNs)~\cite{GCNsKipf2016} have become an increasingly popular deep learning approach for graph-structured data. They offer a powerful tool for learning graph representations, with the ``graph convolution'' operation applying the same linear transformation to all the neighbors of a node followed by a nonlinear activation function. The GCN model is often used as a baseline GNN architecture~\cite{GCNsDeepChen2020}, employing the symmetric-normalized aggregation and self-loop update approach. Overall, GCNs and their variants have proven to be a highly effective and versatile tool for analyzing graph-structured data.

In order to thoroughly examine the graphs created through the \textit{{LineConGraphs}} approach, we make use of a GCN-based model that consists of two graph convolution layers and a Rectified Linear Unit (ReLU) activation layer situated in between. Let us consider a corpus with  containing ``$m$'' utterances, each represented by an $n$-dimensional vector. Using the \textit{LineConGraphs} methodology, we create a graph $G = (V, E)$ where its node feature matrix $X \in R^{[m \times n]}$, we define the adjacency matrix as $A$ and the degree matrix as $D$. The first layer of GCN learns from the n-dimensional feature representation $X^{0}$ to create a new representation $X^{1}$ of dimension ``$k$''. This can be represented as:

\begin{equation}
    X^{1} = GCN(X) = \hat{A} X^{0} W^{0}
\end{equation}
where  $ \hat{A} = D^{1/2} A D^{-1/2} $ is a noramlized adjacency matrix and ``$D$'' indicates normalized degree matrix,   and $W^{0} \in R^{[n \times k]}$ is a weight matrix. The newly extracted representation, $X^{1}$, is then fed into a ReLU activation layer and then into a second GCN layer. The  designed GCN network architecture can be represented as:
\begin{equation}
    Y = argmax( \hat{A} \hspace{0.1cm} ReLU( \hat{A} X W^{0} ) W^{1} )
\end{equation}
where $W^{1} \in R^{[k \times t]}$ is a weight matrix for the second GCN layer and ``$t$'' is the number of labels, i.e., the number of emotions in this case. We then use $argmax$ to select the maximum value among the $t$ values in the final layer as the predicted emotion.

Figure~\ref{fig:gcn-layers-diag} gives an overview of the proposed framework which utilizes the \textit{{LineConGraphs}} with GCN layers for ERC analysis. The GCN model takes the line graph as input and extracts the pretrained model features from EmoBERTa model as node features. Each node represents an utterance, and the sentiment shift between two adjacent utterances is used as edge weight. We use ``$s$'' to represent sentiment shift and ``$ns$'' to indicate no sentiment shift. The feature representation of each utterance after each GCN layer is denoted by $X_{i}^{k}$ where $k$ represents the number of layers. The final layer embeddings are used to recognize the emotion of a given utterance. We evaluate the performance of GCN model for ERC task with and without using sentiment shift as edge weights.

\begin{figure*}[hbtp!]
    \centering
    \includegraphics[width=17cm, height=4.5cm]{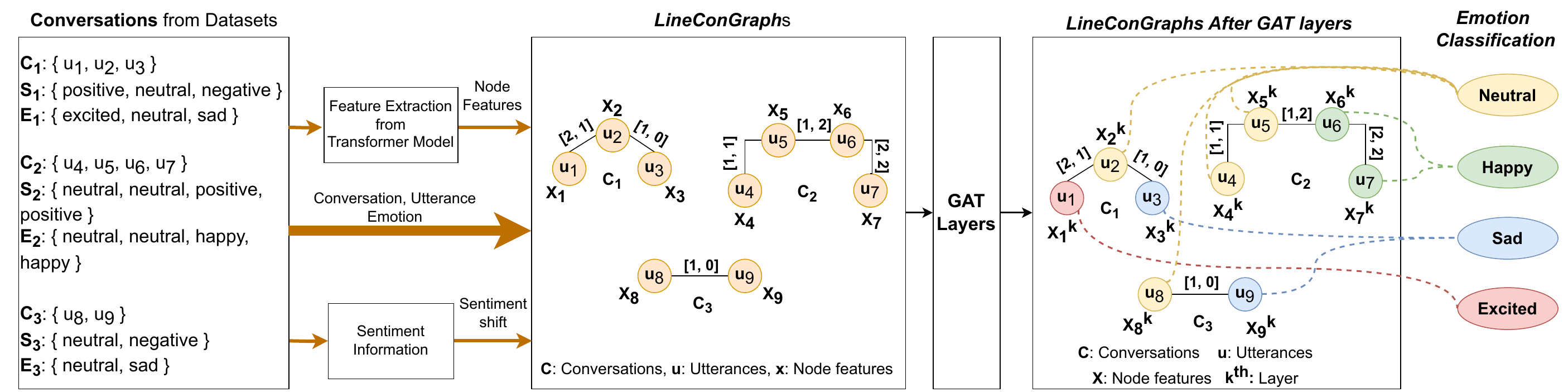}
    \caption{Overview of the proposed framework using the \textit{{LineConGraphs}} with GAT layers for ERC analysis. We use sentiment shift as edge feature   where 0: negative, 1: neutral, 2: positive). [ $u_{i}$: $i^{th}$ utterance ; $X_{i}$: $i^{th}$ node feature extracted for an utterance; $X_{i}^{k}$ indicates feature representation of $i^{th}$ utterance after $k$ GAT layers]}
    \label{fig:gat-layers-diag}
\end{figure*}

%-------------------------------------------------------------------------------------------------------------------------------------

\subsubsection{Graph Attention Networks}

The Graph Attention Networks (GAT) model, proposed by Velivckovic et al. \cite{velivckovic2017graph}, assigns importance or attention weights to each neighboring nodes in a graph. This allows for a neighbor's influence to be considered during the aggregation operation performed over each node's feature representations. However, the original $\textrm{GAT}$ model suffers from the problem of static attention weights. This means that the same attention weight is selected for any set of node representations, limiting the effectiveness of the $\textrm{GAT}$ layer. To solve this issue, a modified version of the $\textrm{GAT}$ layer called $\textrm{GATv2}$ was proposed~\cite{brody2021attentive}, which calculates dynamic attention for any set of node representations. We used the $\textrm{GATv2}$ based model on the constructed conversational graphs using the proposed \textit{LineConGraphs} strategies with node features extracted from transformer based model EmoBERTa to predict the emotion labels associated with each node/utterance. 

For example, a single layer GAT model  learns the new representation $X^{1}_{i}$  for ``$i_{th}$’’  node with an initial node feature $X^{0}_{i}$. This can be formalized as: 
\begin{equation}
\label{gat_layer}
    X^{1}_{i} = \sum_{j \in N(i)} \hspace{0.1cm} \alpha_{i,j} \hspace{0.1cm} W^{0} \hspace{0.1cm} X^{0}_{j}
\end{equation}
where $N(i)$ refers to the set of all neighbour nodes of node $i$; $W^{0} \in R^{[n \times k]}$ refers to shared learnable weight matrix for the first layer and $\alpha_{i,j}$ is the shared learnable attention weight for the node connection from node $i$ to its neighbour node $j$.

The attention weights for a node  $i$ to its neighbor nodes $j$ are calculated as:
\begin{equation} \label{attention_weights_eqn}
\begin{split}
\alpha_{i,j} & = softmax( \hspace{0.1cm} sf (X^{0}_{i}, X^{0}_{j}) \hspace{0.1cm} )\\
       & = \frac{exp \hspace{0.1cm} ( \hspace{0.1cm} sf (X^{0}_{i}, X^{0}_{j}) \hspace{0.1cm} )}{\sum_{j' \in N(i)} exp \hspace{0.1cm} ( \hspace{0.1cm} sf ( X^{0}_{i}, X^{0}_{j'}) \hspace{0.1cm} )}
\end{split}
\end{equation}
where $sf$ is a scoring function that computes a score for every edge  $(i, j)$ and is given as: 
\begin{equation}
\label{scoring_fn}
    sf (X^{0}_{i}, X^{0}_{j}) = \sigma ( a \hspace{0.2cm} [ W^{0} X^{0}_{i} \oplus W^{0} X^{0}_{j} ] )
\end{equation}
where $a$ is a learnable one dimensional vector, $\oplus$ indicates a concatenation operation and $\sigma$ is LeakyReLU activation function.

The softmax function normalizes the scores so that the sum of all attention weights for a node and its neighbors adds up to 1.  

The GAT  learning  model used for ERC analysis is given as follows:
\begin{equation}
\small
%\scalebox{0.8}{
Y = argmax(\sum_{j \in N(i)} \alpha_{i,j}^{1} W^{1} \sigma (\sum_{j \in N(i)} \alpha_{i,j}^{0} W^{0} X^{0}_{j} ) ) 
%}
\end{equation}
where $\sigma$ is ReLU activation function; $W^{1} \in R^{[k \times t]}$ is a weight matrix for the second GAT layer and $t$ is the number of labels, i.e., the number of emotions in this case; $\alpha_{i,j}^{1}$ and $\alpha_{i,j}^{0}$ are attention weights for edge connections between two nodes $i$ and $j$. We then use $argmax$ to select the maximum value among the $t$ final layer values as the predicted emotion label. 
% The designed GAT network produces accurate results in predicting the emotion labels, making it a useful tool for analyzing conversational data.

Figure~\ref{fig:gat-layers-diag} gives an overview of the proposed framework using the \textit{{LineConGraphs}} with GAT layers for ERC analysis. There are two GAT layers used for ERC analysis and as part of the experiments, we embed sentiment shift as edge feature vectors where we directly consider the sentiment labels of connected nodes (0: negative, 1: neutral, 2: positive). Fro example, for conversation $C_3$ in Figure \ref{fig:gat-layers-diag}, we have sentiment labels of utterances $u_8$ and $u_9$ as `neutral' and `negative' respectively. Therefore, the edge feature representation is indicated as $[1, 0]$. The feature representation of each utterance after each GAT layer is denoted by $X_{i}^{k}$ where $k$ represents the number of layers. The final layer embeddings are used to recognize the emotion of an utterance in a given conversation. 
% Please add the following required packages to your document preamble:
% \usepackage{graphicx}

%===================================================================================================================================================================================

\section{Datasets \& Experimental Settings}\label{datasets}

In order to test the effectiveness of our proposed approach, we have utilized two well-known benchmark datasets for emotion recognition - Multi-modal Emotion Lines Dataset $\textrm{MELD}$~\cite{poria2019meld} and interactive emotional dyadic motion capture $\textrm{IEMOCAP}$~\cite{busso2008iemocap}. Both datasets are multi-modal, meaning they include audio, video, and text components. In this paper, we only focused  on the text modality for our experiments. 

%-------------------------------------------------------------------------------------------------------------------------------------

\subsection{MELD} 
It is an extension of the Emotion Lines dataset~\cite{hsu2018emotionlines}, which features conversations from the popular TV series  ``Friends''.  $\textrm{MELD}$ includes $13708$ utterances from  $1433$ conversations, featuring  $304$ unique characters from the show. As shown in Table~\ref{meld-char-table}, this dataset includes seven categorical emotions and sentiment scores for each utterance. The emotional categories and dataset splits used for our analysis are shown in Table~\ref{meld-char-table}. We utilized the training and development splits for fine-tuning our model and the test dataset to evaluate its performance.

% Please add the following required packages to your document preamble:
% \usepackage{graphicx}
%\begin{wraptable}{r}{5.5cm}
\begin{table}[hbtp!]
\centering
\caption{The data statistics  for the  emotional categories  in the  $\textrm{MELD}$ dataset. The number of utterances for each category and the total number of examples used for training, validation and testing purposes.   }
\label{meld-char-table}
\resizebox{0.85\columnwidth}{!}{%
\begin{tabular}{lccc|c}
\hline
\textbf{Emotion}     & \textbf{Train}       & \textbf{Test}        & \textbf{Dev}         & \textbf{Total}       \\ \hline
{Anger}       & 1109                 & 345                  & 153                  & 1607    \\
{Disgust}     & 271                  & 68                   & 22                   & 361    \\
{Fear}        & 268                  & 50                   & 40                   & 358     \\
{Joy}         & 1743                 & 402                  & 163                  & 2308    \\
{Neutral}     & 4710                 & 1256                 & 470                  & 6436    \\
{Sad}     & 683                      & 208                     & 111                  & 1002    \\
{Surprise}    & 1205                 & 281                  & 150                  & 1636    \\\hline
{Total}       & 9989                 & 2610                 & 1109                 & 13708    \\\hline
% \multicolumn{1}{l}{} & \multicolumn{1}{l}{} & \multicolumn{1}{l}{} & \multicolumn{1}{l}{} & \multicolumn{1}{l}{} \\
% \hline
\end{tabular}%
 }

\end{table}
%-------------------------------------------------------------------------------------------------------------------------------------

\subsection{IEMOCAP}
It is a multimodal benchmark dataset for emotion recognition analysis. It includes 12 hours of  data from ten different speakers, divided into 5 sessions with one male and one female speaker each. There are $151$ conversations with $7433$ utterances, covering nine categorical emotions and 3-dimensional labels. For our experiments, we only consider the utterances with at least two annotators agreeing on the emotion label. We focus on six emotions -- happy, excited, neutral, sad, angry, and frustrated, resulting in a total of  $5758$ utterances for model training and $1622$ utterances for model evaluations. Table~\ref{iemocap-char-table} provides data statistics for the $\textrm{IEMOCAP}$ dataset used for our experimental evaluations. We used the first four sessions of the dataset for training and the last session   for model evaluations.

\begin{table*}[hbtp!]
\centering
\tiny
\caption{Statistics of emotional categories in the  $\textrm{IEMOCAP}$ dataset. The number of utterances per emotion category and the total number of utterances for each session.}
\label{iemocap-char-table}
\resizebox{0.85\textwidth}{!}{%
\begin{tabular}{lccccc|c}
\hline
\textbf{Emotion}     & \textbf{Session 1} & \textbf{Session 2} & \textbf{Session 3} & \textbf{Session 4} & \textbf{Session 5} & \textbf{Total} \\
\hline
{Anger}       & 229                & 137                & 240                & 327                & 170                & 1103 \\
{Excited}  & 143                & 210                & 151                & 238                & 299                & 1041 \\
{Frustration} & 280                & 325                & 382                & 481                & 381                & 1848 \\
{Happy}   & 135                & 117                & 135                & 65                 & 143                & 595  \\
{Neutral}     & 384                & 362                & 320                & 258                & 384                & 1708 \\
{Sad}     & 194                & 197                & 305                & 143                & 245                & 1084   \\\hline
{Total}       & 1365               & 1348               & 1533               & 1512               & 1622               & 7380  \\
\hline
\end{tabular}%
}
\end{table*}

%==========================================================================================================
\subsection{Experimental Settings}

In our experiments, we use PyTorch Geometric package~\cite{pytorchGeometric} for implementing our proposed approach. We used a total of two GCN and two GAT layers, respectively, along with categorical cross-entropy as the loss function. Both models employed an AdamW optimizer with a learning rate of $10^{-3}$  and a weight decay parameter of $10^{-4}$. We employed sentiment shift information as edge weight in the case of GCN for MELD, where the value of $s$ is ``1'' indicating no change in sentiment between connected nodes, and value of $ns$ is ``-1'' which indicates a change in sentiment. Similarly, for the $\textrm{IEMOCAP}$ dataset, we used ``1'' to indicate no change and ``2'' to indicate a change in sentiment labels\footnote{The parameters $s$ and $ns$ were determined empirically.}. Since the  $\textrm{IEMOCAP}$ dataset does not include sentiment scores, we used a RoBERTa-based sentiment classification model~\cite{loureiro2022timelms} to obtain sentiment labels and generate sentiment shift edge weights for the GCN model. In the GAT model, we encode the sentiment shift as an edge feature which contains the sentiment labels of connected nodes. We also construct a fully connected graph where every node is connected to every other node in a conversation and evaluated both GCN and GAT models on this graph using the same parameter settings as the line graph.

We used predefined train, development, and test datasets for experimental evaluations on the  $\textrm{MELD}$ dataset. As for the $\textrm{IEMOCAP}$ dataset, we follow the state-of-the-art approaches, MMGCN~\cite{hu2021mmgcn}, ConGCN~\cite{ConGCNZhang2019} and DialogGCN~\cite{ghosal2019dialoguegcn}, and trained the model on the first four sessions, while evaluating it on the fifth session. This evaluation ensures that our proposed method is evaluated in speaker-independent mode. To assess the performance of the proposed system, we report a weighted average $F1$ score and compute confusion matrices to visualize how the system performed within and across different emotion categories.

%===============================================================================================================

\section{Results and Discussions}\label{results}

In this section, we compare our proposed approach against the state-of-the-art approaches for ERC. We also perform a comparative analysis of speaker-dependent and independent models. Additionally, we analyze the performance of GCN and GAT models with and without sentiment shift. Finally, we visualize how emotions can often overlap and confuse with other categories of emotions highlighting the importance of accurately detecting and distinguishing between them. 
%Furthermore, we perform  extensive experimentation to compare with a Fully Connected Graph construction approach using GCN and GAT models.
 
\subsection{Comparison with State of the Art}

We conducted extensive experimentation to assess the efficacy of the proposed \textit{LineConGraphs} for ERC analysis and compared it with the current SOTA methods. In Table~\ref{comparison-table-graphs}, we have presented a detailed performance comparison of our proposed line graph approach with GCN and GAT models against the current SOTA ERC task for text modality on the $\textrm{IEMOCAP}$ and $\textrm{MELD}$ benchmark datasets. Our line graph approach with GCN (\textit{LineConGCN}) has surpassed all other methods with an impressive $68.18\%$ on the  $\textrm{MELD}$ dataset, showcasing a significant $2.5\%$ improvement over previous methods. Additionally, incorporating the sentiment shift information as edge weights (\textit{LineConGCN}ss) has led to an outstanding performance of $74.37\%$. This represents a significant $8.57\%$ improvement over prior methods. The line graph approach with GAT  (\textit{LineConGAT}) further improved the ERC performance with an F1 score of $76.50\%$. This showcases a substantial $10.70\%$ improvement over previous methods. Similar outcomes have been observed with sentiment shift as edge features. Overall, we can conclude that the GCN and GAT models built on top of \textit{LineConGraphs} have achieved state-of-the-art results on the  $\textrm{MELD}$ dataset.

Our line graph approach with GAT model (\textit{LineConGAT}) demonstrated a comparable performance to previous methods on the  $\textrm{IEMOCAP}$ dataset achieving an impressive F1 score of $64.58\%$. However, we found that adding sentiment shifts as an edge feature did not lead to any significant improvement in performance. This could be due to the lack of ground truth sentiment scores available for the dataset, even though we used the RoBERTa based model to generate sentiment labels. 

As indicated in the Table~\ref{comparison-table-graphs}, our proposed approach is compared only to state-of-the-art methods that are consistent in experimental settings and test sets used for model evaluations. However, in the following section, we provide additional comparisons with other methods and discuss inconsistencies in model evaluations (Table~\ref{spkr_embed}).

%We didn't make a direct comparison between our approach and other models like GraphMFT, EmoBERTa, and RGCN due to variations in the test sets and experimental settings. However, we provided more information on our experimental setups and a comparison with our proposed method in Table~\ref{spkr_embed} in the following section.

\begin{table}[hbtp!]
\center
\normalsize
\caption{Performance comparison of proposed \textit{LineConGraphs} with prior methods. Abbreviations:\textit{LineConGCN} – \textit{LineConGraphs} with GCN model; \textit{LineConGCN}ss – GCN model built on \textit{LineConGraphs}  with sentiment shift as edge weights; \textit{LineConGAT} – GAT model built on \textit{LineConGraphs}; \textit{LineConGAT}ss – GAT model built on \textit{LineConGraphs} with sentiment
shift as edge features; WA-weighted average F1 score in percentage. Note: \textbf{Bold} indicates the best performing model; -- indicates values not reported for the dataset.}
\label{comparison-table-graphs}
\renewcommand{\arraystretch}{1.2}

\resizebox{0.48\textwidth}{!}{%
%\fontsize{10pt}{10pt}\selectfont
\begin{tabular}{c c c}
\hline
{\multirow{2}{*}{\textbf{Approach}}}      & \textbf{IEMOCAP}  & \textbf{MELD }\\\cline{2-3} 
                                          & {\textbf{WA F1 (\%)}} & {\textbf{ WA F1 (\%) } }  \\ \hline 
DialogueRNN \cite{majumder2019dialoguernn}            & 62.75        & -- \\   
DialogGCN \cite{ghosal2019dialoguegcn}                & 64.18        & 58.10 \\  
ConGCN \cite{ConGCNZhang2019}                    & --             & 57.4  \\ 
% ConGCN \cite{ConGCNZhang2019} (audio)                   & --             & 42.2  \\ 
% ConGCN \cite{ConGCNZhang2019} (multi)                   & --             & 59.4  \\ 
GAT + spkr \cite{StaticDynamicSaxena2022} &--            &65.8\\
% MMGCN \cite{hu2021mmgcn} (audio)                                 & 54.66        & 52.63 \\ 
MMGCN \cite{hu2021mmgcn}                                 & 62.35        & 57.72 \\ 
SKIER \cite{SkierWei2023}                               & --              & 67.39       \\
\textit{ICON} \cite{ICONPoria2018}                      &63.5               &\\
\textit{DIMMN} \cite{DIMMNWen2023}                      & 58.6       & 64.1                 \\
% MMGCN \cite{hu2021mmgcn} (multi)                                & 66.22        & 58.65 \\ 
% GraphMFT \cite{JiangLi2023Neuro} (multi)                       &68.07          &58.37\\
% RGCN+edgeloss \cite{RGCNChoi2021}                      &65.08          &55.98\\
% EmoBERTa  \cite{kim2021emoberta}                      &68.57          &65.61\\
\textit{\textit{LineConGCN}} (ours)                               & 63.89      &68.18\\
\textit{\textit{LineConGCN}}$_{ss}$ (ours)                         & 60.62  & 74.37 \\ 
% \textit{\textit{FCConGraphs}} + GAT                               & 59.44        & 74.43 \\
{\textit{\textit{LineConGAT}} (ours)  }                          & {\bf 64.58}      & \textbf{76.50} \\
\textit{\textit{LineConGAT}}$_{ss}$  (ours)                          & 63.52        & 76.28 \\ \hline
\end{tabular}%
}

\end{table}

%While we didn't compare our approach directly to other models such as GraphMFT, EmoBERTa, and RGCN because of differences in test sets and experimental settings, refer to Table~\ref{spkr_embed} in the following section for more information on our experimental setups and comparison with our proposed method.

\begin{table*}[h!]
\centering
\caption{  $\textrm{MELD}$ dataset: prior methods that are dependent on speaker information for ERC analysis. ~~Abbreviations: w/ spkr: with speaker embedding, w/o spkr: without speaker embedding, m: multimodal and t: only text modality.  rs: random split ( 8:2 ), s: session wise evaluations, \textit{LineConGAT}: line conversation graph with GAT model.  Note: {\bf Bold} indicates our proposed model and session wise (s) or random split (rs) evaluation is only applicable to IEMOCAP dataset.}
\label{spkr_embed}
\renewcommand{\arraystretch}{1.2}
 \resizebox{0.75\textwidth}{!}{%
\begin{tabular}{lcccc}
\hline
\multicolumn{1}{c}{\multirow{2}{*}{\textbf{Approach}}}      & \multicolumn{2}{c}{\textbf{ MELD}}    & \multicolumn{2}{c}{\textbf{ IEMOCAP}}    \\ \cline{2-5} 
\multicolumn{1}{c}{}          & \multicolumn{1}{r}{\textbf{w/ spkr }} & \multicolumn{1}{r}{\textbf{w/o spkr  }}   & \multicolumn{1}{r}{\textbf{w/ spkr }} & \multicolumn{1}{r}{\textbf{w/o spkr  }} \\ \hline  
\textit{MMGCN} \cite{hu2021mmgcn} (m) (s)                  &58.65             & 58.38              & 66.22           & 65.76  \\
\textit{ConGCN} \cite{ConGCNZhang2019} (m) (s)             & 59.4             & 57.4               & --              & --   \\
 \textit{ConGCN} \cite{ConGCNZhang2019} (t) (s)             & 57.4             & 55.3               & --              & --  \\
\textit{DialogRNN} \cite{majumder2019dialoguernn} (m) (s)   & --               & --                 &57.38            & 55.56 \\
\textit{EmoBERTa} \cite{kim2021emoberta} (t) (rs)            & 65.61            & 65.07              &68.57           & 64.02 \\
\textit{GraphMFT} \cite{JiangLi2023Neuro} (m) (rs)          & 58.37            & 58.04              &68.07            & 65.61 \\
\textit{RGCN+edge loss} \cite{RGCNChoi2021} (t) (rs)          & --            & 55.98              &--                & 65.08 \\
\textit{SCCL} \cite{ClusterContrastiveYang2023} (m) (rs) &--             & --        & --    & 69.81       \\
 \textit{LineConGAT} (Ours) (t) (s)                          &--                & {\bf 76.50}        &--           & {\bf 64.58}\\
\hline                                   
\end{tabular}%
 }

\end{table*}

%-------------------------------------------------------------------------------------------------------------------------------------
\subsection{Speaker Dependence in ERC}
\subsubsection{ $\textrm{MELD}$}
According to Table~\ref{spkr_embed}, various multimodal and unimodal approaches incorporated speaker information for ERC analysis. However, these approaches may not be compelling in determining the emotion of the utterance in a specific context. It is important to note that speakers can express their emotions differently over time, making speaker-dependent models impractical for new speakers in real-time settings. Interestingly, except for ConGCN, all other methods achieved similar performance with and without speaker embedding. Moreover, speaker embedding did improve the model's performance for the  $\textrm{IEMOCAP}$ dataset, but it may be due to the limited number of speakers present in the dataset( only ten speakers). However, for the  $\textrm{MELD}$ dataset, there are  $260$ speakers, and there is no significant difference in performance with or without speaker embedding. Therefore, the proposed \textit{LineConGraphs} approach outperformed prior methods that relied on speaker embeddings and multiple modalities for ERC analysis. The best part is that our proposed model does not rely on speaker information, making it a practical and effective ERC application for new and unknown speakers in real-world contexts. To further improve its performance, we plan to use speech and video embeddings as node features in our line graph representation.

%-------------------------------------------------------------------------------------------------------------------------------------

\subsubsection{ $\textrm{IEMOCAP}$: Inconsistencies}

Based on a thorough review of prior studies on the $\textrm{IEMOCAP}$ dataset, it seems that there are inconsistencies in the experimental settings used to evaluate the performance of models in terms of ER analysis. However, Padi et al.~\cite{saralaimprovedER2021,saralamultimodalER2022} have addressed these inconsistencies and have assessed the model's performance for both unimodal (text and audio) and multimodal scenarios for ER analysis. On the other hand, previous methods used to measure the performance of models for ERC analysis on the IEMOCAP dataset have used varying settings, which could lead to inaccurate results. Therefore, it is crucial to ensure that these settings are consistent to accurately evaluate the model's performance. In an effort to perform a fair comparison with the SOTA methods, we have compared our proposed approach with the current SOTA methods and their respective settings used for ERC analysis in Table~\ref{spkr_embed}.

Notably, MMGCN, GraphMFT, RGCN, and SCCL surpassed our model performance for the  $\textrm{IEMOCAP}$ dataset. However, GraphMFT and SCCL employed random data splits of 80\% of data for training and 20\% for testing their model performance for ERC analysis. On the other hand, EmoBERTa reported model performance on 10\% of data, while 90\% of data was used for model training. Moreover, MMGCN and GraphMFT models are multimodal, where they trained their models on audio, text, and video modalities. In contrast, our model is evaluated solely on text modality and in a speaker-independent manner. We trained our model in the first four sessions with eight speakers, four male, and four female, and evaluated it in the last session, which includes two speakers, one male and one female. This setup is consistent with DialogGCN, ConGCN, DialogRNN, and MMGCN models. While GraphMFT, RGCN, and EmoBERTa compared their model performances with DialogGCN, ConGCN, and DialogRNN models, it is not equitable to compare because the latter models were evaluated in session-wise, which is a speaker-independent way and represents around 20\% of the $\textrm{IEMOCAP}$ dataset, whereas, the former models measured their model performances on a random split which is the speaker-dependent way with either 20\% or 10\% data.

\subsection{Fully Connected Conversation Graphs}
Section~\ref{related_work}, we provided an extensive literature review on various graph construction strategies for conversational modeling. Most previous methods focused on constructing fully connected graphs where every utterance in a conversation connects to every other in the given conversation. However, we believe this method could provide too much context, which may impact the model's ability to learn subtle nuances such as emotion shifting and mirroring. To assess the effectiveness of the proposed \textit{LineConGraphs} approach with GCN and GAT models, a baseline called Fully Connected Conversation Graphs (\textit{FCConGraphs}) was developed. This baseline approach will help us understand the effectiveness of our proposed method.

\begin{table}[h!]
\centering
\caption{Performance comparison of proposed line graph approach with fully connected graph on the  $\textrm{IEMOCAP}$ and  $\textrm{MELD}$ datasets.~~Abbreviations: \textit{FCConGCN}: fully connected conversation graph with GCN model, \textit{FCConGCN}$_{ss}$: \textit{FCConGCN} with sentiment edge weights, \textit{FCConGAT}: fully connected conversation graph with GAT model, \textit{FCConGAT}$_{ss}$: \textit{FCConGAT} with sentiment edge features.}
\label{fc-results-table}
\renewcommand{\arraystretch}{1.3}
 \resizebox{0.45\textwidth}{!}{%
\begin{tabular}{lcc}
\hline
\multicolumn{1}{c}{\multirow{2}{*}{\textbf{Approach}}}      & {\textbf{ IEMOCAP}}    & {\textbf{MELD}}    \\ \cline{2-3} 
                                            &{\textbf{WA F1 (\%)}} &{\textbf{WA F1 (\%)}} \\ \hline
\textit{FCConGCN}                              &51.70                    & 46.84      \\
\textit{FCConGCN}$_{ss}$                       & 52.09                   & 48.93   \\
\textit{\textit{LineConGCN}}                    &63.89                   &68.18 \\ 
\textit{\textit{LineConGCN}}$_{ss}$             & {60.62}                & {74.37} \\ \hline
\textit{FCConGAT}                             & 63.97                    & 74.43       \\
\textit{FCConGAT}$_{ss}$                     & 62.38                     & 74.43      \\
\textbf{\textit{LineConGAT}}                 & \textbf{64.58}             & \textbf{76.50}  \\
\textit{LineConGAT}$_{ss}$           & 63.52            & 76.28 \\
\hline                                    
\end{tabular} }
\end{table}

We perform experiments by constructing a fully connected graph with GCN and GAT models. We also used sentiment information as edge weights for GCN and edge features for GAT. As shown in Table~\ref{fc-results-table}, we compared the performance between fully connected and line graph approaches. Interestingly, we found that the line graph with the GCN model (LineConGCN) outperformed the fully connected graph with the GCN model (FullyConvGCN), and similarly, the line graph with the GAT model outperformed the FullyConvGAT model. This indicates that the line graph is the most effective approach for ERC analysis. Furthermore, the long-term context is aggregated by adding more GCN or GAT layers by allowing the information to be passed from past and future nodes in the graph without the need for explicit input. Therefore, the context is implicitly derived from the proposed \textit{LineConGraphs} with the number of layers used in the model.

\subsubsection{Sentiment Shift}
Based on our analysis (Table~\ref{fc-results-table}), we found that the addition of sentiment shift information in the form of edge weights and features does have a positive impact on the performance of the \textit{LineConGraphs} strategy when applied to the  $\textrm{MELD}$ dataset. However, we do observe a decrease in the performance of the GCN model for the  $\textrm{IEMOCAP}$ dataset, which we suspect could be due to incorrect sentiment labeling. Despite this, we observed that the GAT model was able to compensate for this misclassification to some extent due to its ability to learn the attention weights dynamically. We conclude that while adding sentiment shift information can be beneficial in some cases, the GAT model's ability to learn the importance of connected edges is superior to the provision of edge features.

%-------------------------------------------------------------------------------------------------------------------------------------
%====================================================================================
 \begin{figure*}[bhpt!]
 \begin{center}
 %\fbox{\rule{0pt}{1in}
 \subfloat[\small { $\textrm{MELD}$ }]{\includegraphics[width=8cm, height=6cm,trim=0cm 0cm 1.80cm 0cm, clip=true]{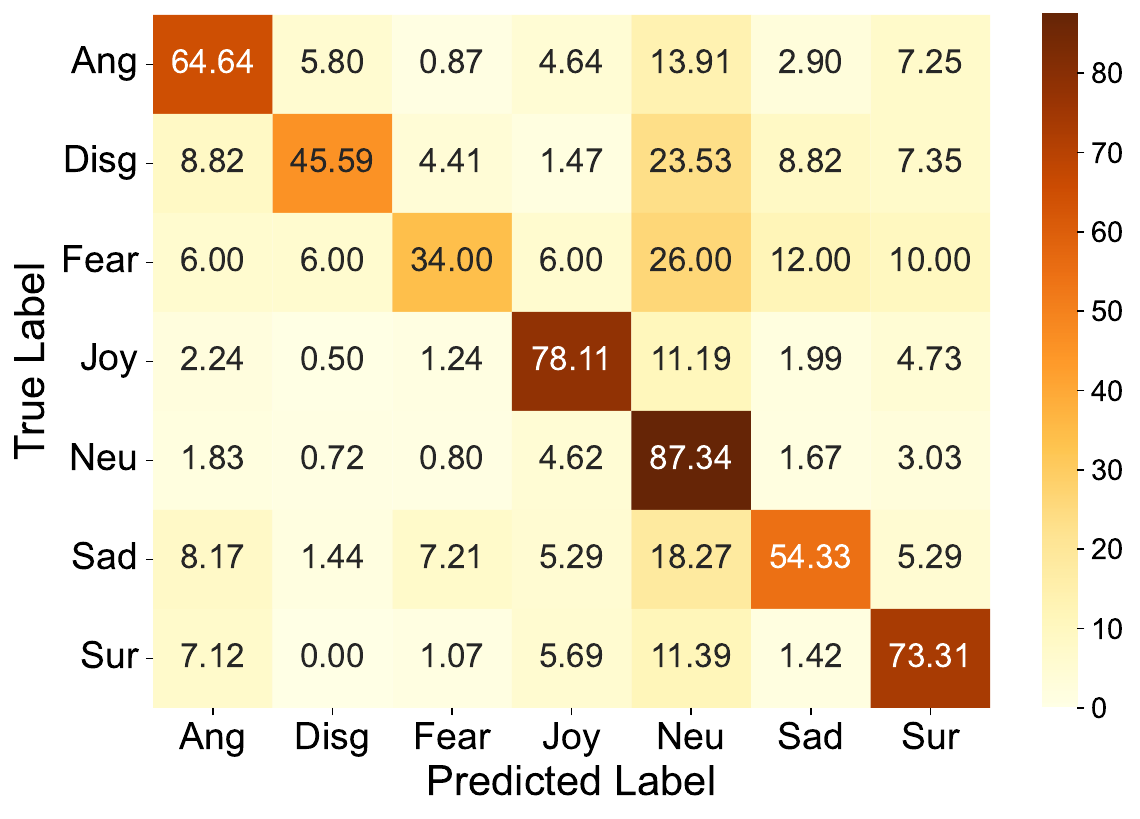}}
 \hspace{0.2cm}
  \subfloat[\small { $\textrm{IEMOCAP}$ }]{\includegraphics[width=8cm, height=6cm,trim=0.8cm 0cm 1.80cm 0cm, clip=true]{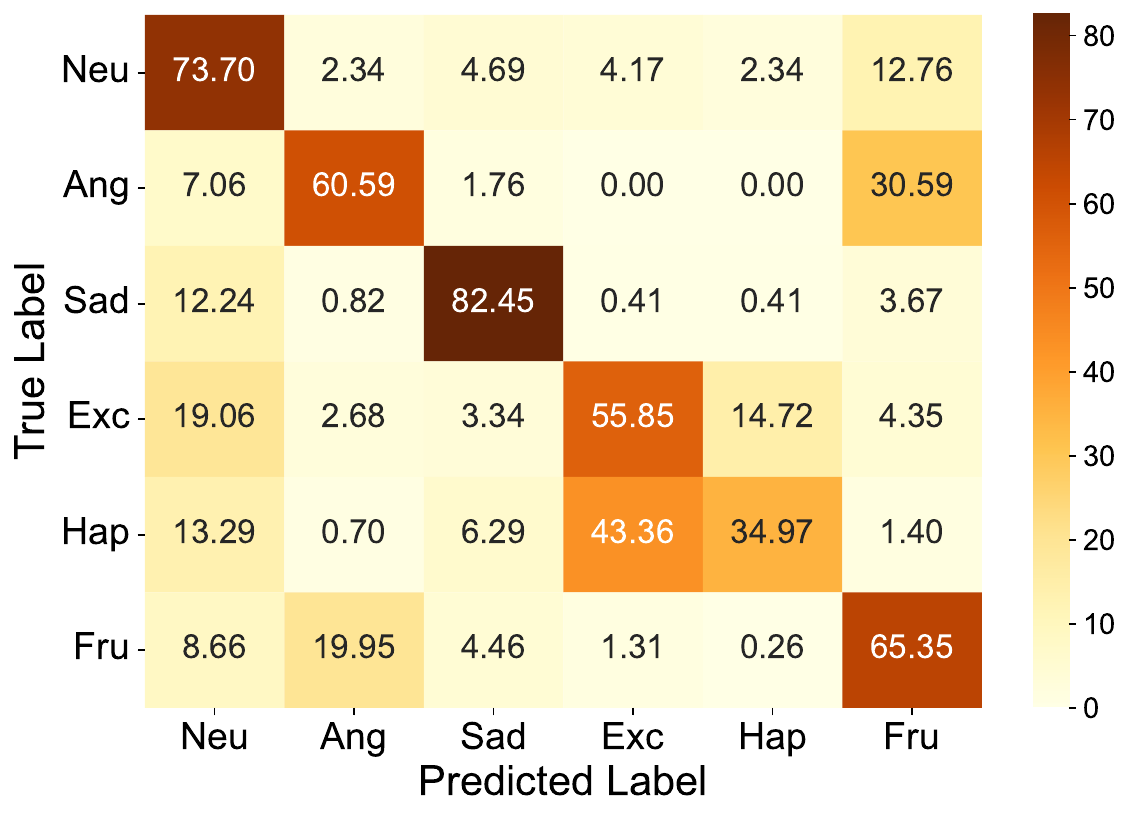}}\\
     %}
 \end{center}
 %\vspace{-0.2cm}
    \caption{ Confusion matrices of \textit{LineConGAT} model for  $\textrm{MELD}$ and  $\textrm{IEMOCAP}$ datasets. Abbreviations: Ang -Angry, Disg - Disgust, Neu - Neutral, Sur - Surprise, Exc - Excited, Hap - Happy, Fru - Frustration; Note: we reported the emotion-wise accuracies for test sets. }
 \label{fig:confusion}
 \end{figure*}

%====================================================================================
 \begin{figure*}[bhpt!]
 \begin{center}
 %\fbox{\rule{0pt}{1in}
 \subfloat[\small { $\textrm{MELD: LineConGCNs }$ }]{\includegraphics[width=8cm, height=6cm,trim=0cm 0cm 1.80cm 0cm, clip=true]{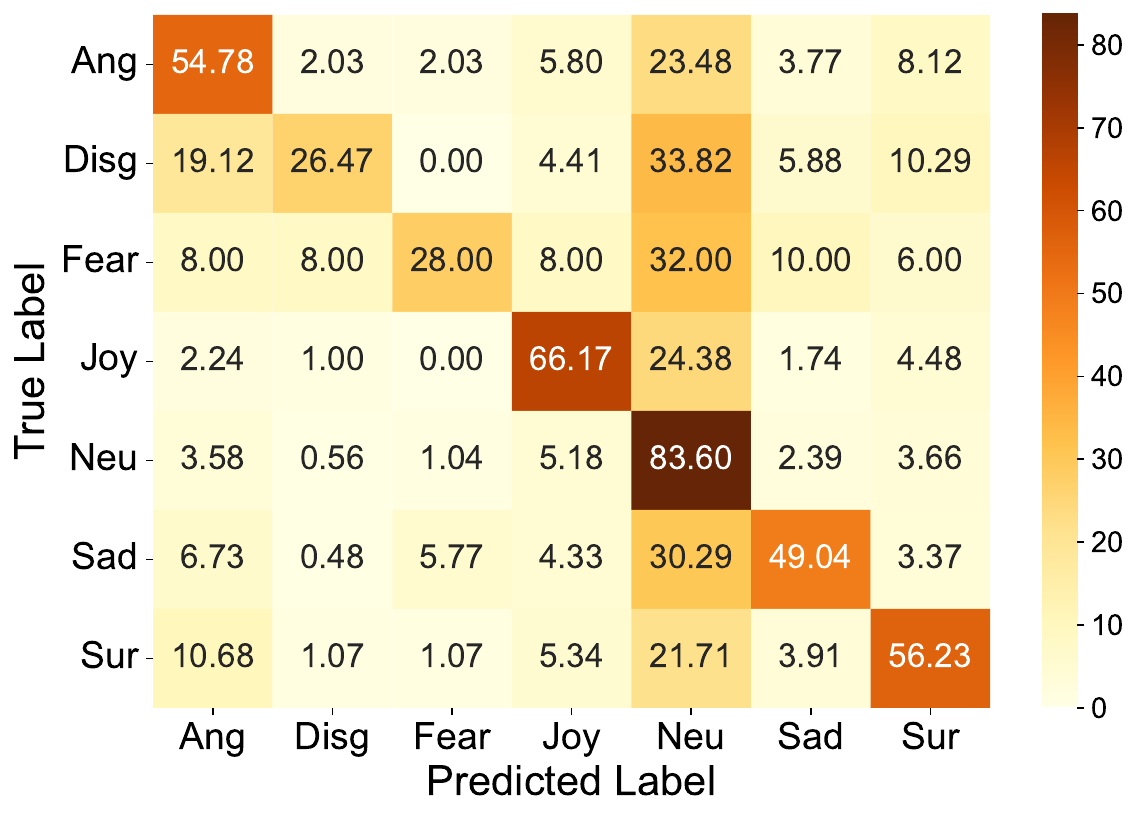}}
  \hspace{0.2cm}
  \subfloat[\small { $\textrm{MELD: LineConGCNs with sentiment edge weights }$ }]{\includegraphics[width=8cm, height=6cm,trim=0.8cm 0cm 1.80cm 0cm, clip=true]{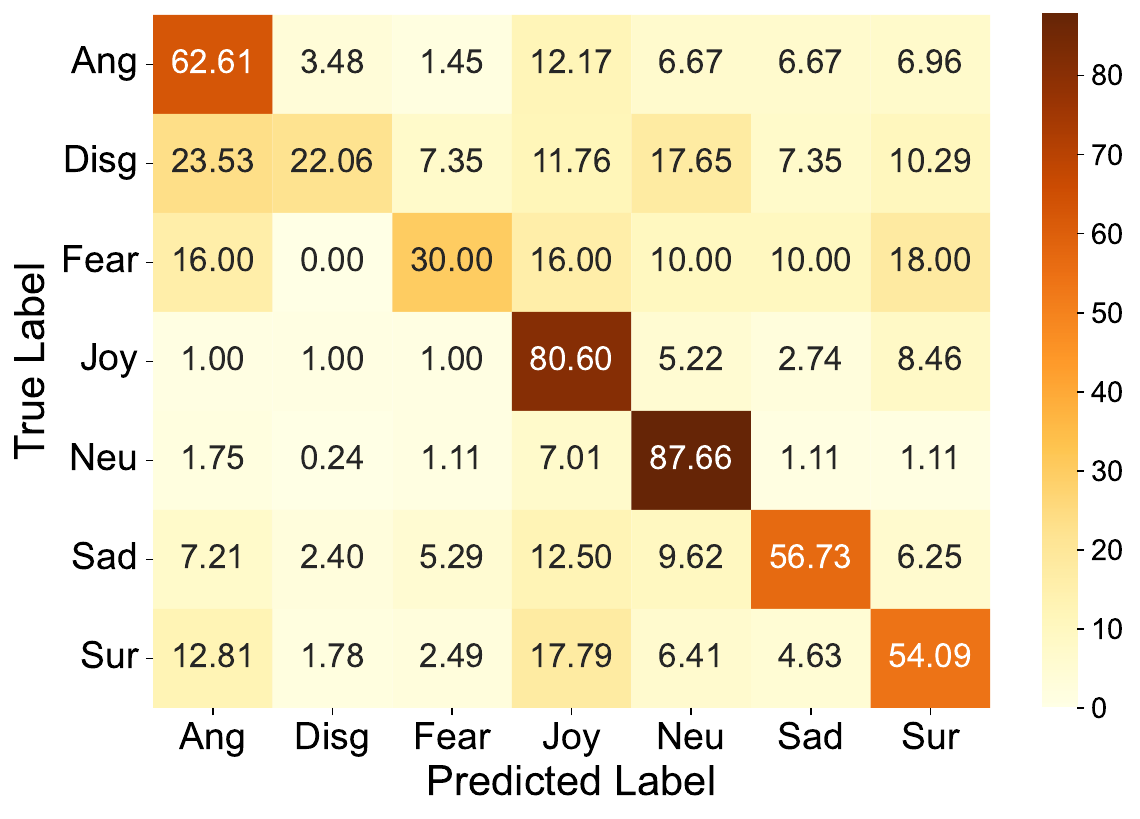}} \\
 % \subfloat[\small { $\textrm{IEMOCAP: LineConGCNs}$ }]{\includegraphics[width=8cm, height=6cm,trim=0cm 0cm 1.80cm 0cm, clip=true]{images/confusion_matrix_IEMOCAP_GCN.pdf}}\\

 %\hspace{0.2cm}
 % \subfloat[\small { $\textrm{IEMOCAP: LineConGCNs with sentiment edge weights}$ }]{\includegraphics[width=8cm, height=6cm,trim=0.8cm 0cm 1.80cm 0cm, clip=true]{images/confusion_matrix_IEMOCAP_GCN_sent_edge_weights.pdf}}
     %}
 \end{center}
 %\vspace{-0.2cm}
    \caption{ Confusion matrices of \textit{LineConGCN} models and  \textit{\textit{LineConGCN}}$_{ss}$ for $\textrm{MELD}$ dataset. Abbreviations: Ang -Angry, Disg - Disgust, Neu - Neutral, Sur - Surprise. Note: we reported the emotion-wise accuracies for test set. }
 \label{fig:confusion1}
 \end{figure*}

%================================================================================
\subsection{Error Analysis}

To illustrate the performance of the proposed line graph approach within and across different emotion categories, we report confusion matrices.  Figure~\ref{fig:confusion} shows the confusion matrices of the best performing models (\textit{LineConGAT}) for the two benchmark datasets. The \textit{LineConGAT} model confuses the ``disgust'', ``fear'', and ``sadness'' categories of emotions with the ``neutral'' class quite often while performing the best on the ``neutral'', ``joy'', and ``surprise'' emotions. 

Similarly, in the case of  $\textrm{IEMOCAP}$ dataset, the \textit{LineConGAT} model exhibits the highest performance in predicting the ``sadness'' and  ``neutral'' categories of emotions. It should be noted that  ``anger'' emotion is frequently confused with ``frustration'', ``excited'' emotion is often confused with ``neutral'' and ``happiness'' and the ``happy'' category of emotion is often confused with ``neutral'', and ``excited''. For the $\textrm{IEMOCAP}$ dataset, the most similar emotions are getting confused, compared to the $\textrm{MELD}$ dataset, where misclassifications are with the ``neutral’’ emotion category. 

%-------------------------------------------------------------------------------------------------------------------------------------

As per the results presented in Table~\ref{fc-results-table}, integrating sentiment shift as an edge feature to the \textit{LineConGCN} model has improved the ERC performance for the MELD dataset. The confusion matrices were plotted to visually compare the \textit{LineConGCN} and \textit{LineConGCN} with sentiment edge weights (\textit{\textit{LineConGCN}}$_{ss}$). As shown in Figure~\ref{fig:confusion1}(a), the \textit{LineConGCN} model confuses the ``Disgust'' category of emotion with ``neutral'' 33\% of the time, and with ``angry'' 19\% of the time. Similarly, ``surprised'' is confused with ``neutral'' 21\% and with ``angry'' 10\% of times. On average, the ``neutral'' emotion category got confused with the six categories of emotions for 27\% of the time. In contrast, in \textit{\textit{LineConGCN}}$_{ss}$ (Figure~\ref{fig:confusion1} (b)), the ``Disgust'' category of emotion is mostly confused with ``anger'' 23\% of the time and with ``neutral'' 17\% of the time. Also, ``surprise'' is most confused with ``joy'' 17\% of the time and with ``anger'' 12\% of the time. The overall confusability of the ``neutral'' emotion category is reduced from 27\% to 9\% in \textit{\textit{LineConGCN}}$_{ss}$ compared to \textit{LineConGCN}. It is worth noting that \textit{\textit{LineConGCN}}$_{ss}$ confuses between more similar emotions, while \textit{LineConGCN} mostly confuses with the ``neutral'' emotion category. The study concluded that incorporating sentiment shift into graph modeling significantly improved the ERC performance by reducing the misclassification rate with the ``neutral'' emotion category. In the future, we intend to generate the sentiment labels for the IEMOCAP dataset using large language models like ChatGPT, and study the impact of the sentiment shift feature on ERC analysis.

%which is similar to the LineConGAT model./ confusability between ``surprise'' and ``joy'', ``disgust'' and ``anger'' is more convincing as compared to being misclassified as a ``neutral'' emotion. The study concluded that integrating sentiment shift into graph modeling significantly improves the ERC performance by reducing the misclassification rate with ``neutral'' emotion. In the future, we plan to generate the sentiment labels for the IEMOCAP dataset using ChatGPT and analyze the sentiment shift feature impact for ERC analysis.

%===============================================================================================================
\section{Conclusion \& Future Scope}\label{conclusion}

In this paper, we proposed a novel  \textit{LineConGraphs} for representing the utterances using a short context of preceding and succeeding utterances in a given conversation for ERC analysis. We evaluated the effectiveness of our proposed line graph representation by utilizing GCN and GAT-based models and tested them on two benchmark datasets, $\textrm{IEMOCAP}$ and  $\textrm{MELD}$. Our experiments have shown that the line graph with the GAT model outperformed all other models, while the line graph with sentiment weights using the GAT model ranks second over all previous methods. Our findings also indicate that the incorporation of sentiment shift improved the ERC performance of the GCN model. However, the inclusion of sentiment shift as an edge feature did not benefit the GAT model. This suggests that attention weights can effectively identify the crucial subtleties in the given utterances for ERC analysis. Our approach is noteworthy for being speaker-independent, utilizing sentiment shift, and achieving high ERC performance by solely utilizing text conversations and their contexts. Therefore, it is an ideal ERC solution that can be deployed in real-world scenarios, even for speakers unknown to the model. To further enhance the ERC performance and address similar emotion misclassification issues, we plan to investigate the integration of other modalities, such as vision and audio. Furthermore,  we intend to explore other graph construction strategies in order to extend the capabilities of \textit{LineConGraphs} and make them an even more robust and dependable solution for ERC.

\section*{Acknowledgments}
The authors want to thank Omid Sadjadi and Michael Majurski for their helpful comments and suggestions to make the paper even better.

\section*{Disclaimer}
Certain equipment, instruments, software, or materials are identified in this paper in order to specify the experimental procedure adequately.  Such identification is not intended to imply recommendation or endorsement of any product or service by NIST, nor is it intended to imply that the materials or equipment identified are necessarily the best available for the purpose. 

%Commercial products are identified in this document  are specify the experimental procedure adequately. Such identification is not intended to imply recommendation or endorsement by NIST, nor is it intended to imply that the products identified are necessarily the best available for the purpose.

\bibliographystyle{IEEEtranN}
\bibliography{custom}

\end{document}